# A Feasibility Experiment on the Application of Predictive Coding to Instant Messaging Corpora

Thanasis Schoinas
Legal Technology Solutions
Navigant Consulting Europe Ltd.
London, United Kingdom
e-mail: thanasis.schoinas@navigant.com

Ghulam Qadir
(former) Legal Technology Solutions
Navigant Consulting Europe Ltd.
London, United Kingdom

*Abstract* - **Predictive coding, the term used in the legal industry for document classification using machine learning, presents additional challenges when the dataset comprises instant messages, due to their informal nature and smaller sizes. In this paper, we exploit a data management workflow to group messages into day chats, followed by feature selection and a logistic regression classifier to provide an economically feasible predictive coding solution. We also improve the solution's baseline model performance by dimensionality reduction, with focus on quantitative features. We test our methodology on an Instant Bloomberg dataset, rich in quantitative information. In parallel, we provide an example of the cost savings of our approach.**

## I. INTRODUCTION

From the view of text analytics, instant messaging (IM) can be considered a synchronous, less structured and shorter version of email, in turn seen as a condensed version of a document. In many cases, IM is just a casual means of communication with no contextual link to other artefacts, nevertheless related to a subject of interest or matter of investigation. With the growing volume of instant message communications in both desktop and mobile devices, as well as its institutionalised forms, such as Instant Bloomberg, there is a growing interest in the legal and electronic discovery community in ways to preserve and analyse these communication artefacts, as well as a recognition of the additional challenges they present, as compared to emails and documents. It is also recognised that advanced analytics, such as Technology Assisted Review (i.e. Predictive Coding) can help legal teams "weed out crucial information faster and with greater accuracy, yet the complex metadata and other peculiarities of chat data have previously confounded such tools" [1].

Coding IMs with regards to their responsiveness is one area where these challenges appear. The fine granularity of chat messages makes assigning a responsiveness or privilege code in its native form - per chat line - impractical.

Conversations may be left open for days and span different topics during their lifecycle, during which participants may join and leave at any time, increasing complexity of custodian assignment.

For the above reasons, it makes sense grouping individual chat lines into collections of text, which can be coded and assigned to custodians, thus making a human review economically feasible. In an author identification study, the fine granularity challenge was overcome by splitting or combining chat logs into segments of certain number of characters [6]. In our study, we adopted a time-dependent approach, grouping messages by conversation and time, with day chat collection documents being a realistic option, on the assumption that instant communication about a topic - especially a non-casual one - is highly likely to conclude within the day it was initiated.

Additional challenges arise when we consider technology assisted review, aimed at prioritizing documents that are more likely to be responsive, based on knowledge acquired by the model training examples. IM text usually contains a higher number of abbreviations and spelling mistakes, thus increasing the vocabulary size (i.e. dimensionality) of the corpus. The data sparsity resulting from small text sizes and the less explicit and descriptive nature of text, has an impact on the effectiveness of classification algorithms, which has a negative impact on potential review savings and consequently cost effectiveness of the TAR exercise. A higher number of numeric and special or technical domain-specific characters further increase that complexity. For these reasons, a text pre-processing approach focusing on token standardisation and extraction of semantic value, initially in the form of rule-based tagging, can enhance effectiveness of such efforts. We take a step towards that direction with the treatment of numbers, in a corpus rich in quantitative information such as Instant Bloomberg (IB).

We used well-established supervised learning algorithms for feature selection and classification, preceded by two additional components for message grouping and features engineering. Simulating an active learning scenario with limited number (5,000) of Instant Bloomberg training documents, the baseline model used the "original" day chat texts, with precision @80% recall in the order of 60.7%.

Following our proposed quantitative features treatment, this precision reached 65%. An experiment was also carried out using a much larger set of training data, with similar improvements and precision reaching 74%.

Our study is work in progress, but results so far demonstrate that addressing IMs' fine granularity can make predictive coding a commercially - from a review costs perspective - viable solution, whereas reducing the higher dimensionality caused by various kinds of quantitative information (digits) can lead to model performance gains, which we are directly translating to review cost savings.

## II. Background - Related Work

Predictive coding is a use case of binary classification of documents with respect to their coding (Responsive/Not responsive, Privileged / Not privileged). Various classification algorithms have been used in various packages offered by eDiscovery software and service providers. Research into the various algorithms and packages offered for classification is out of scope of this study. Details about the properties and various options (Logistic Regression vs Support Vector Machines, normalised frequency vs *tf*tf*idf* etc.) for the settings of the predictive coding environment we are using are contained in [2].

Some of the focus areas of advanced statistical and machine learning techniques for processing of IMs are creation of conversational agents (chatbots) [5] or, in the public security domain, author identification based on token, syntactic and structural features [6], [7]. We could not find any documented example of using those techniques for predictive coding of instant messages in the legal domain.

In the context of the model improvement part of our study, there is substantial literature on the analogy between classification of text and that of protein sequences [3] in which we can identify similarities between deployed classification approaches. More specifically and as a motivation for our approach, Thomas et al. [4] have provided evidence of successful use of aminoacid higher level properties in classification of T-Cell (immune system cell) receptor sequences. Their approach is based on a bag-of-words algorithm, utilising the frequencies of p-tuples (a/acid duplets, triplets and so on) as features, in an analogy to words (or bigrams, trigrams and so on) in language processing. Vectors representing each amino acid by some of its physicochemical properties, such as polarity and molecular volume, are also added. In this biological paradigm, the method follows from research providing indications that local features of protein sequence may contain hidden information that is "aminoacid-agnostic", as it refers to the property rather than the aminoacid itself. Even though the above paradigm cannot be directly applied to our natural language domain, we could nevertheless recognise similarities, so we transferred elements of the approach into exploiting properties of the tokens making up a day chat document. The only higher-level (semantic) property we have exploited so far is that of a token being of quantitative type, i.e. we assumed that the reviewer would be interested to know that a number (whether it be monetary amount, interest rate or of any other nature) exists in the message but not interested in the value itself.

## III. Method

### A. Processing environment components and settings

We implemented two pre-processing steps aiming at standardising the document content:

- A chat deduplication and compression workflow; this consolidates and de-duplicates messages from various participants' individual chat logs, also accounting for conversation timelines. The output of this step is the two datasets, comprising raw and normalised "day chat" documents.

- A text pattern-based numerical tagging component that focuses on features engineering, in the form of text processing prior to model training. By tagging in this context we refer to the task of entity extraction from text, which in our case focuses on identifying numerical tokens, annotating them as such and then replacing them with their annotation.

Subsequently, logistic regression models were created and tested, according to optimal settings identified empirically. Those were features of token normalised frequency, no stemming, minimum token length 3 and "top 20000" tokens. Prior to training the model with LR, an information gain assessment was performed, based on which the tokens are ranked, so the "top N" can be subsequently selected for model training.

### B. Dataset

The ~67k coded documents were stored in a Relativity document repository, with granularity at the consolidated "day chat" level. The dataset included feeds from Instant Bloomberg, the core chatroom of interest, as well as other IM sources, which were used for results comparison.

#### 1) Raw versus normalised text

The differences between original (raw) and normalised day chat texts are:

- Removal of "Participant X joined/left the room"

- Removal of "N hour and NN minutes since previous line"

- Removal of "Participant X Says" phrases

In summary, the normalisation process creates anonymised versions of the messages by eliminating conversation participants; in addition, it removes information about chat silence periods (Fig. 1). The normalisation step has only been performed on Instant Bloomberg and not on other chat sources in our set.

Results were evaluated for both message versions, to assess the effect of participant information in the text.

### 2) Removing and Tagging numerical tokens

During the initial exploratory models, it was observed that a substantial number of numeric tokens were being assigned high logistic regression coefficients, indicating that the model might be overfitting by learning from potentially noise features. With the aim to reduce noise in the wider context of dimensionality reduction, one option would be to remove all numbers from model training. However, the resulting loss of information about frequency of numeric tokens in a message was not desirable, plus this approach would not treat digits that were part of a longer string.

The alternative of converting all numbers (sequences of consecutive digits) into one universal "*[NUM]*" tag was therefore implemented as a pre-processing step.

Finally, a second tag was tried along with the first one, by separating timestamps out of all tokens tagged as numeric and assigning a new *[TIMESTAMP]*" tag to those. Fig. 1 demonstrates the main text variants used.

### 3) Batch Training

For a batch training scenario, we used a validation set comprising a random sample of 14% (~7.3k) of the entire IB dataset at our disposal (~51k documents) with the remaining 86% used for training. Note that our "Full" dataset still contained documents whose size would drop to very low levels after normalisation, but the raw document versions - which are the focus of this paper - had a size of over 0.3KB.

Figure 1. Normalisation and tagging on day-chat text. "[NUM]" and "[TIMESTAMP]" tags are shown in the corresponding versions

### 4) Iterative Training

To simulate reality as much as possible, we also tried a set of models ("5k") with only 5k training documents and the remaining (~29k) population used as validation set, resembling the real-life scenario of an iterative approach (e.g. active learning), which additionally assumes no prior keyword knowledge, i.e. the 5k documents are a random sample. Further to that, the dataset was filtered to include only documents whose extracted text size would be over 0.3KB both in the raw and the normalised version, in contrast to the batch ones, where only the raw version met this size criterion. This filtering was done to ensure that the dataset comprised documents that contained reasonable amount of information for the first step of an active learning experiment.

An inverted version ("29k") of the above sets was also used, i.e. ~29k training and 5k validation, to assess the effect of increasing training set size.

### C. Results

### 1) Model Performance

Our main finding is the jumps in performance from not-tagged to tagged versions. Results for precision at 80% recall are shown in TABLE I.

One can also observe a small drop in predictive performance due to the use of normalised text, in the order of 1% precision @80% recall, and a similar loss across most recall levels (not shown). This can be attributed to the loss of participants information in the normalised text version.

Results in the form of ROC curve for the best performing model, Raw 5k single-tag, are shown in Fig. 2 and indicate that the tagged model performs better over most recall (true positive rate) levels.

For a single-metric model comparison, we computed approximate values of the Areas Under the Receiver Operating Characteristic Curve, shown in Fig. 4. As typically expected, performance improves substantially with training size.

In terms of the effect of our enhancements, an intuitive interpretation is that, as the training set becomes larger, the effect of dimensionality reduction is not as profound.

TABLE I.      PRECISION @ 80% RECALL COMPARISONS (UNTAGGED, ONE AND TWO TAGS)

| | Precision @ 80% recall | | |
|---|---|---|---|
| | *Untagged* | *"[NUM]" Tag* | *"[NUM]" and "[TIMESTAMP]" Tags* |
| *Raw - 5k* | 60.72% | 65.03% | 65.08% |
| *Norm - 5k* | 59.82% | 63.40% | Not tested |
| *Raw - 29k* | 72.66% | 74.07% | 73.98% |
| *Raw - Full* | 71.83% | 73.21% | 72.67% |
| *Norm - Full* | 69.00% | 72.21% | Not tested |

This is also indicated by the reduced differences between the non-tagged and tagged model Precision/Recall curves, plotted for three training sizes in Fig. 3. Based on these plots, it can also be argued that the gains are more persistent at lower to medium recall levels, i.e. higher cutoff scores, which are typically used for active learning exercises. A more detailed study, as well as a statistical significance test for the differences in the AUROC curves would be useful for a full estimate of the effect of our enhancements and the recall regions where these are potentially stronger, as what we are observing in Fig. 3 could be just random fluctuations due to the different validation sets. For the time being we can continue with our intuitive interpretation and resort to the conclusion that the gain from our model improvements is more evident - hence persistent for larger training sets - at higher cutoff scores, where positive examples can be better separated from negative ones.

Finally, training a model with text containing the two tags, resulted into marginally better results, compared to single tag, in the case of the 5k runs and marginally worse results for the large training set (Fig. 4). Albeit small, the area increases in line with the number of tags indicate that there is potential to adding further tags for smaller training sets - such as in an active learning scenario - but, as before, a more detailed study is needed to assess this.

### 2) Features Importance

Feature importance of the *"[NUM]"* tag in the Full raw single-tag model is very low, as it does not even make it to the first 20,000 features considered by logistic regression. When two tags are used, its information gain is elevated and it is assigned a moderately positive coefficient, whereas the second tag *"[TIMESTAMP]"* is in the top 70 positive tokens.

Conversely in the 5k raw runs, the *"[NUM]"* tag seems to be the second strongest contributor to the non-responsive class for single tagging, while in the two-tag model its negative coefficient is halved, whereas the *"[TIMESTAMP]"* tag is now also assigned a moderately negative coefficient. For attribute importance purposes, the 5k runs can be dismissed as less descriptive, due to the small training sets. To further support the latter statement, it is worth mentioning that the number of tokens in the 5k models does not exceed 12,000.

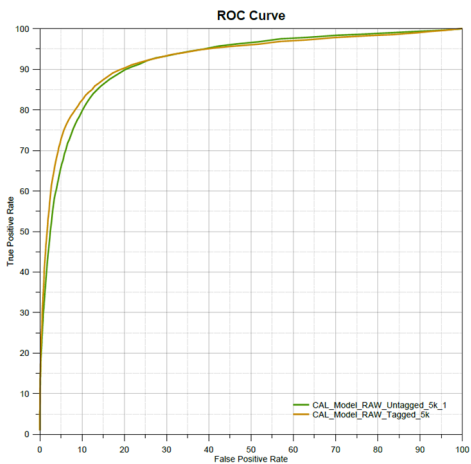

Figure 2.  Effect of single tagging on ROC curve, Raw 5k models

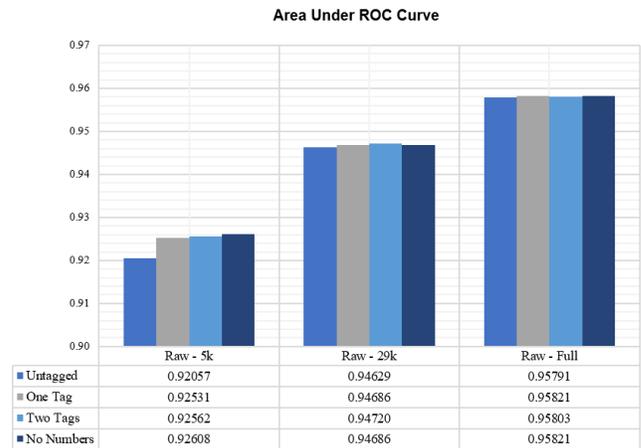

Figure 4.  Area Under ROC curve (trapezoidal approximation on 100 intervals)

| | Raw - 5k | Raw - 29k | Raw - Full |
|---|---|---|---|
| Untagged | 0.92057 | 0.94629 | 0.95791 |
| One Tag | 0.92531 | 0.94686 | 0.95821 |
| Two Tags | 0.92562 | 0.94720 | 0.95803 |
| No Numbers | 0.92608 | 0.94686 | 0.95821 |

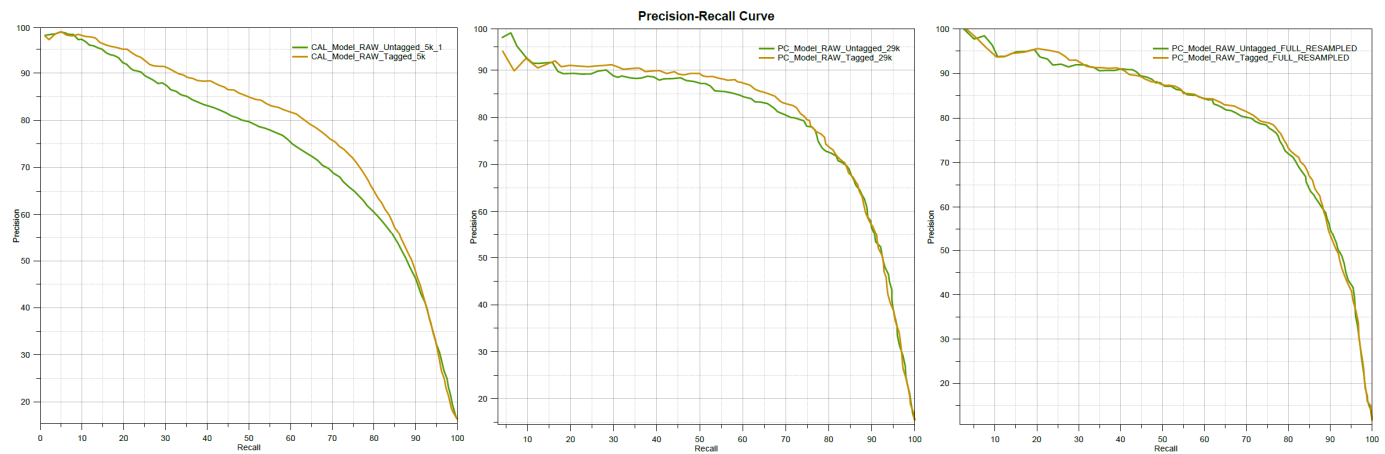

Figure 3.  Effect of single tagging on Precision/Recall, for increasing training set size 5k > 29k > Full

Considering the more reliable Full models, given that our tags are only two features out of 20,000, we cannot claim a strong indication that the more numbers a message contained, the more likely it would be to be responsive, as this was only the case when both tags are used - and not in the case of the single-tag model. It is also observed that apart from the top 10 positive terms, which had significantly higher coefficients (TABLE II. ), weights were spread closely among a large number of both positive and negative tokens. In addition, the top weighted tokens seem to be terms that one would expect to have low normalised frequencies, in contrast to our two tags, for which we can observe high normalised frequencies; this may be the reason why the tags were not assigned high weights and is in line with text analytics literature, in that text classification involves learning with many relevant features [8]. Considering the above, we can argue that the observed increase in performance is more likely to be attributed to the decrease in dimensionality caused by eliminating numbers, rather than the tags' predictive power itself.

The fact that it is dimensionality reduction rather than information gain that led to the increase in performance is validated by a comparison of results between our best performing single-tag model and a model of the same text, but this time with the actual tags themselves removed. This is equivalent to removing all instances of digits in the original text. The latter model yielded almost identical results with our single-tag best performing model, as shown in Fig. 5.

TABLE II.    TOP FEATURES FOR THE BASELINE AND THE BEST PERFORMING 5K MODEL

| Baseline | | Raw 5k Best | |
|---|---|---|---|
| *Token* | *LR model coefficient* | *Token* | *LR model coefficient* |
| Isdafix | 12.55 | isdafix | 14.46 |
| isdafix2 | 10.29 | done | 6.45 |
| isdafix4 | 10.28 | fix | 5.10 |
| Done | 5.15 | isda | 4.40 |
| <Intercept> | -5.05 | axed | 2.94 |
| fix4 | 4.87 | confm | 2.93 |
| Isda | 4.68 | conf | 2.91 |
| fix2 | 3.64 | dael | 2.79 |
| axed | 3.46 | confirm | 2.74 |
| Fix | 3.37 | isdafx | 2.36 |
| confm | 3.17 | bro | 2.32 |
| *(redacted)@(redacted)* | 3.09 | the | -2.23 |
| *(redacted)@(redacted)* | -3.03 | adjust | 2.18 |
| *(redacted)@(redacted)* | 2.61 | off | -2.09 |
| Adjust | 2.57 | guan | 1.91 |
| Conf | 2.49 | agreed | 1.80 |
| Dael | 2.34 | swed | 1.78 |
| Bro | 2.30 | chf | -1.73 |
| Axes | 2.13 | heap | 1.71 |
| confirm | 2.13 | pattern | 1.71 |
| *(redacted)@(redacted)* | 2.02 | screen | 1.70 |
| Heap | 2.00 | confo | 1.69 |
| Austin | 1.97 | ddb | 1.66 |
| *(redacted)@(redacted)* | 1.94 | shaun | -1.66 |
| Isdafx | 1.94 | thnks | 1.62 |

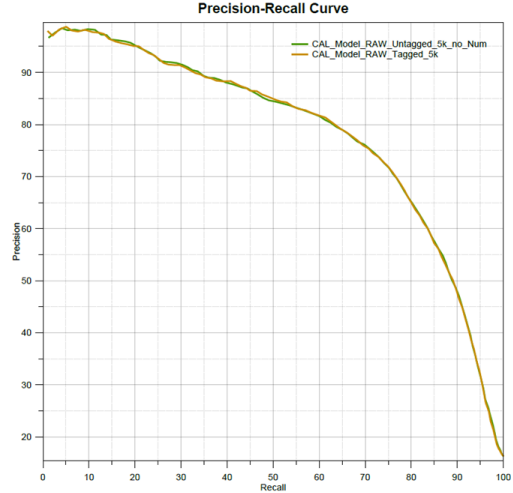

Figure 5.    (Single) tagged versus complete numeric elimination with no tagging, Raw 5k models

### D. Anticipated business benefit

By way of putting the performance we have achieved so far into a review cost savings context, the baseline precision in our raw 5k dataset (60.72% @ 80% recall) can be interpreted as:

Assume we have a trained model with 5k Instant Bloomberg day chats and a new corpus of 100,000 day chats is received for coding (this would be a far higher number of actual chat lines, possibly in the millions, prior to our compression workflow).

To retrieve 80% (13,008) of the responsive chats - assuming the new corpus has the same responsiveness rate as our validation set, that is 16.26% - we would have to review 21,423 day chats, i.e. 21.42% of the 100,000 documents.

Taking this scenario one step further, the 4.31% precision gain (rather, 7.1% on initial precision basis – from 60.72% to 65.03%) we have achieved with the elimination of numbers in the raw text, translates to the same percentage gain in the number of scored documents that would require manual review to achieve the 80% desired recall level. In figures, for the population of 100,000 new day chats, we would now need to review 20,003 instead of 21,423 documents, that is a further 1,420 fewer documents than with the baseline model.

## IV.    CONCLUSIONS

By grouping instant messages through a compression and deduplication workflow, we have generated day chat documents that can be reviewed and coded by legal professionals in a cost-effective manner and produce an information-rich and class-separable dataset, in contrast to a dataset at a chat line level. Subsequently, we utilise feature selection and classification to perform predictive coding of these day chat messages, which further reduces the costs of review.

Finally, we propose an additional dimensionality reduction stage, which yields noticeable increase in model performance and can provide the basis for even further improvements, by reducing noise quantitative tokens and replacing them with their characterisation, in a first step towards named entity recognition.

Our pipeline is tested using both a traditional batch (full training set) and an incremental (e.g. active learning) example. Our review savings calculations suggest that it can be used in both scenarios.

## V. FUTURE WORK

Further evaluations, as well as refinement of tags will be needed to assess the approach's full potential. We are currently examining tagging numbers per their semantic value (e.g. amount, date, reference, rate etc.); this will require a regular expression matching approach and will enable addition of some semantic meaning to the data points, e.g. we are interested to know that a monetary amount exists somewhere in the message but not interested in the amount per se. This can later also help derive features as combinations of tags, e.g. an <amount> followed by a <rate> might be correlated with responsive documents. In the same context, a model with minimum token length 1 was also tried, so as to not lose standalone, currency or other, symbols but performance deteriorated; it may therefore make sense to convert such symbols to three letter tokens (e.g. "$" to "USD") for these to be considered by the model, without having to resort to minimum token length of 1, which introduces further noise.

In addition, considering tags in the form of [Low/Medium/High length of silence], [Low/Medium/High chat word count] will potentially help. These should be added as text inside each document.

By the definition of logistic regression, tokens are treated independently. This means that a logistic regression model will not be able to detect "cliques" of participants appearing in responsive messages, unless we introduce tokens to represent groups of persons. As it stands, each participant is weighted individually, as is the case with every other token.

Training and assessing models on each chat source separately, aims to avoid confusing the modelling engine with potentially different vocabularies used at each room. This may impact generalisation, but can be addressed by applying the same segregation during scoring. The approach could broadly be compared to using keyword search to narrow down the documents used for model training, thus allowing the model to focus on the vocabulary used in those documents that contain the desired keywords. Our results following this principle and applying the process on IB data only are inconclusive so far and a richer dataset would be required for such a test.

Finally, in an ideal world of unlimited project resources, chat text could do with cleansing, as it contains many abbreviations, quite often with spelling mistakes. We may benefit by reviewing and rationalising the vocabulary used before running predictive coding. It might be beneficial to process the corpus vocabulary with a string matching (e.g. Levenshtein distance) or other method to cluster similar words. This can further reduce dimensionality and improve accuracy.

## ACKNOWLEDGMENT

Many thanks to the co-author, Ghulam Qadir for his substantial contribution to features engineering - creating the day chats and integrating tags to the baseline text. We would also like to thank Richard Chalk for his encouragement and for providing the support and resources needed to complete this study, as well as his guidance on the data pre-processing and management pipeline. Special thanks to Katie Jensen for her business and technical guidance, Jon Fowler for introducing us to the problem, Nathaniel Huber-Fliflet for advising on document content and structure and finally Jianping Zhang for his expert insights. Thanks also to Jack Bullen and Meghan Ryan for helping with data pre-processing and model training runs.